\documentclass[letterpaper, 10 pt, conference]{ieeeconf}  
\usepackage[utf8]{inputenc}

\IEEEoverridecommandlockouts                              

\usepackage{cite}
\usepackage{amsmath,amssymb,amsfonts}
\usepackage{algorithmic}
\usepackage{graphicx}
\usepackage{textcomp}
\usepackage{xcolor}
\usepackage{bm}
\usepackage{balance}
\usepackage{multirow}
\usepackage{makecell}
\usepackage{booktabs}
\usepackage{threeparttable}
\usepackage{pifont}
\usepackage{soul}
\usepackage[pagewise]{lineno}
\usepackage{comment}
\usepackage{cuted}
\usepackage{booktabs}
\usepackage{multirow}
\usepackage{graphicx}
\usepackage[hidelinks]{hyperref}

\hypersetup{
colorlinks=false,
pdfborder={0 0 0}
}

\usepackage{xcolor}
\definecolor{projectred}{RGB}{185,75,75}

\soulregister{\cite}7
\soulregister{\ref}7
\soulregister{\ding}7
\usepackage[ruled,linesnumbered]{algorithm2e}

\title{\LARGE \bf BifrostUMI: Bridging Robot-Free Demonstrations and Humanoid Whole-Body Manipulation}

\author{Hongwu Wang\textsuperscript{$\dagger$}, Chenhao Yu\textsuperscript{$\dagger$},  Youhao Hu\textsuperscript{$\dagger$}, Jiachen Zhang, Yuanyuan Li and Shaqi Luo\textsuperscript{*}\\
\textit{Beijing Academy of Artificial Intelligence}
\thanks{$\dagger$ Equal Contribution}
\thanks{
* Send all correspondence to sqluo@baai.ac.cn}
}

\begin{document}
\maketitle
\vspace{-1.2em}

\begin{strip}
    \vspace{-0.8em}
    \centering

    \includegraphics[width=0.98\textwidth]{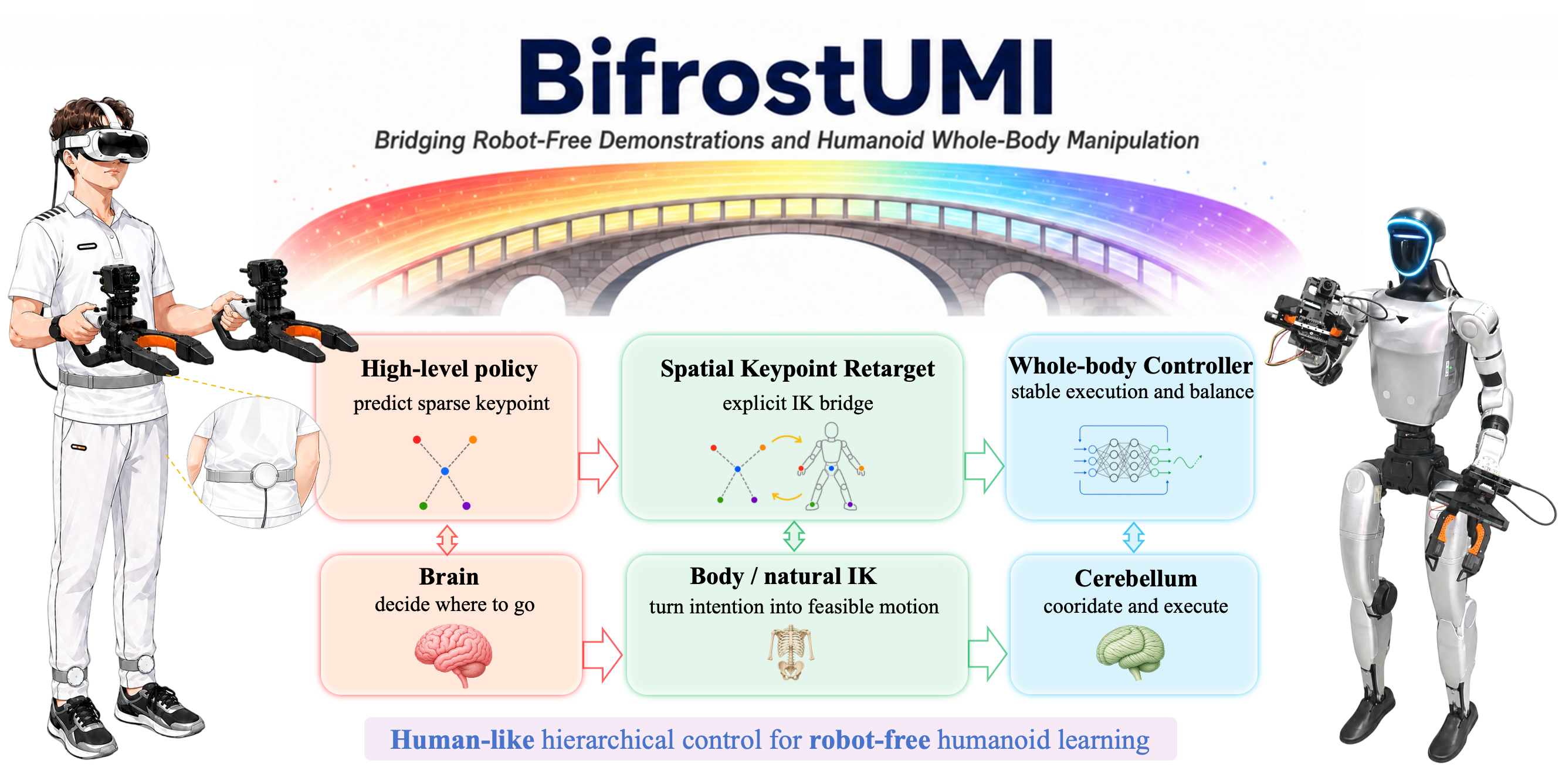}

    \vspace{-0.8em}

    \refstepcounter{figure}
    \label{fig:bifrost_overview}

    \begin{minipage}{0.95\textwidth}
    \footnotesize
    \textbf{Fig.~\thefigure. Overview of BifrostUMI.}
    BifrostUMI provides a robot-free data collection and learning framework for humanoid whole-body skills. Human demonstrations are collected with portable VR--UMI devices and represented as sparse task-relevant spatial keypoints. A high-level policy predicts future keypoint motions and gripper actions, a Spatial Keypoint Retargeting module converts them into feasible humanoid whole-body references, and a low-level whole-body controller executes the retargeted motion with balance and stability. This hierarchical design bridges natural human demonstrations and deployable humanoid whole-body behaviors.
    \end{minipage}

    \vspace{-1.2em}
\end{strip}

\vspace{-0.8em}

\thispagestyle{empty}
\pagestyle{empty}

\begin{abstract}
High-quality demonstration data are essential for humanoid robot skill learning, especially for whole-body behaviors that require coordinated perception, locomotion, and manipulation. Existing data-collection methods largely rely on robot teleoperation, which is constrained by hardware accessibility, operator expertise, and limited efficiency. Inspired by the Universal Manipulation Interface (UMI), we propose \textbf{BifrostUMI}, a portable and robot-free framework for humanoid whole-body data collection. BifrostUMI uses lightweight VR devices and UMI-inspired grippers to collect sparse human keypoint trajectories, wrist-view observations, and gripper actions. These demonstrations train a high-level policy to predict future keypoints, which are retargeted to robot-native whole-body references and executed by a whole-body controller. Experiments in five real-world scenarios demonstrate the effectiveness of the proposed framework and validate the collected demonstrations for transferable humanoid whole-body skill learning. 
\noindent\textbf{Project page:}
\href{https://baai-aether.github.io/BifrostUMI/}
{\textcolor{projectred}{baai-aether.github.io/BifrostUMI}}

\end{abstract}

\begin{keywords}
Humanoid Robot, Robot-Free Data collection, Whole-Body Manipulation
\end{keywords}

\section{Introduction}

Data have become a central substrate for embodied intelligence~\cite{droid, twist2}. For humanoid robots, acquiring task-relevant demonstrations remains challenging, especially for whole-body skills that require coordinated manipulation, posture adjustment, and locomotion. Most existing systems rely on robot-in-the-loop teleoperation, where human operators directly control the physical humanoid to generate training trajectories~\cite{twist2,humanplus,clone,mobiletv,dreamtouch,h2p,omnih2o}. Although effective, teleoperation requires robot access, skilled operators, safety supervision, and repeated hardware operation, making large-scale data collection costly and inefficient.

Robot-free human demonstration offers a promising alternative. The UMI series of works~\cite{umi, omniumi,scalinglaw,umift, activeumi} shows that portable handheld devices and wrist-centric visual observations can replace conventional teleoperation for robot-arm manipulation, enabling efficient collection of in-the-wild demonstrations. Similar ideas have been extended to quadruped and aerial robots~\cite{umionleg,umionair}, and HoMMI explores UMI-style collection for dual-arm mobile manipulation~\cite{hommi}. However, extending this paradigm to humanoid whole-body manipulation is nontrivial. Unlike robot arms, humanoids must coordinate manipulation with balance, posture adjustment, stepping, and morphology-dependent whole-body motion. HuMI investigated humanoid data collection with body tracking and UMI-style grippers~\cite{humi}, but the problem of building a portable, low-cost, robot-free pipeline that can produce deployable whole-body humanoid policies remains underexplored.

In this work, we propose \textbf{BifrostUMI}, a portable robot-free data collection and learning framework for humanoid whole-body manipulation. Instead of collecting demonstrations through the target robot, BifrostUMI captures human whole-body intent using a PICO 4 VR system and UMI-inspired handheld grippers. The system records synchronized wrist-view images, gripper widths, and a compact five-keypoint representation consisting of the pelvis, two gripper TCPs, and two feet. This sparse spatial representation preserves the task-relevant spatial structure of the demonstrated whole-body motion and provides an explicit interface for human-to-humanoid retargeting.

To deploy robot-free demonstrations on a humanoid robot, BifrostUMI adopts a hierarchical framework that maps human demonstration intent to executable whole-body motion. A high-level visuomotor policy predicts sparse spatial keypoint motions and gripper actions, a Spatial Keypoint Retargeting module converts them into robot-native whole-body references, and a learned whole-body controller executes the retargeted motion on the physical humanoid. This design provides an explicit bridge between robot-free human demonstrations and coordinated humanoid whole-body execution.

We validate BifrostUMI on a Unitree G1 humanoid robot across five real-world tasks covering single-arm manipulation, bimanual coordination, dynamic ball throwing, whole-body bending, and locomotion-manipulation. The results show that BifrostUMI can transform robot-free human demonstrations into deployable humanoid skills, while collecting valid demonstrations more efficiently than a teleoperation baseline.

Our main contributions are threefold:
\begin{enumerate}
    \item We introduce \textbf{BifrostUMI}, a portable and low-cost VR--UMI system for robot-free humanoid whole-body demonstration collection.
    \item We propose a \textbf{sparse spatial keypoint retargeting} interface that preserves task-relevant geometric structure for human-to-humanoid whole-body transfer.
    \item We design a \textbf{hierarchical policy framework} for learning humanoid whole-body skills from robot-free demonstrations, and empirically verify that the collected data can be transformed into deployable real-world policies.
\end{enumerate}

\begin{figure*}[t]
    \centering
    \includegraphics[width=0.8\textwidth]{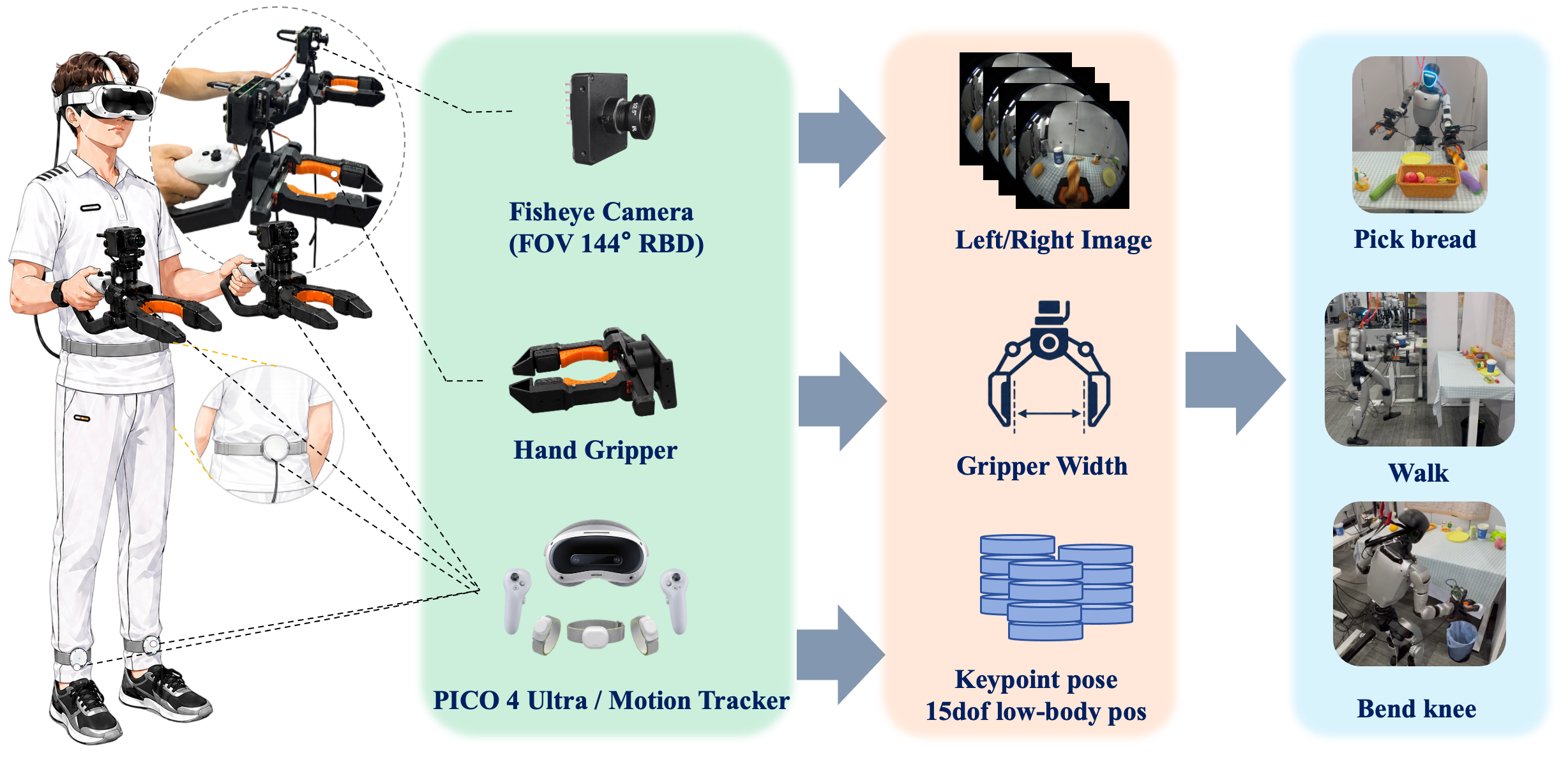}
    \caption{\textbf{BifrostUMI Data Acquisition System.}
The data acquisition platform consists of a PICO-based motion capture setup, including two foot-mounted trackers and one waist-mounted tracker, together with two instrumented grippers, each equipped with a fisheye camera. The system synchronously records multimodal observations, including wrist-view images from the fisheye cameras, human keypoint states obtained via the PICO SDK, and gripper aperture measurements derived from motor encoder readings. These heterogeneous data streams are jointly used to train a high-level policy, which is subsequently deployed for real-time control of robot motion.}
    \label{data}
\end{figure*}
\section{Related Works}

\subsection{Humanoid Data Collection}

Recent advances in humanoid learning have motivated various data-collection frameworks.
Robot-in-the-loop teleoperation methods collect demonstrations by directly controlling the target platform~\cite{aloha,maloha,clone,dreamtouch}.
For humanoids, systems such as TWIST2, CLONE, and Touch Dreaming extend this paradigm to whole-body or humanoid manipulation settings.
Although these methods provide embodiment-consistent trajectories, they typically require access to the physical robot, skilled operators, careful safety supervision, and repeated hardware operation, making large-scale data collection costly and inefficient.

Another line of work explores robot-free or human-centric demonstration collection.
EgoHumanoid~\cite{egohumanoid} studies egocentric human demonstration and human-to-humanoid transfer, but such methods must explicitly bridge the embodiment gap between human motion and humanoid-executable actions, which remains challenging for precise whole-body manipulation.

UMI-style interfaces offer a portable and economical alternative for manipulation data collection~\cite{umi}.
HoMMI extends this idea to dual-arm mobile manipulation~\cite{hommi}, but does not target full-body humanoid control.
HuMI further demonstrates the feasibility of UMI-style data collection for humanoids~\cite{humi}; however, its Vive-based tracking system introduces additional hardware cost and calibration complexity, and its retargeting procedure is tightly coupled with the downstream whole-body controller.

In contrast, BifrostUMI provides a portable, low-cost, and robot-free framework for whole-body humanoid data collection by combining VR-based full-body sensing with UMI-inspired handheld grippers.

\subsection{Whole-Body Visuomotor Policies}

Recent work has also explored whole-body visuomotor policy learning for humanoid robots.
TWIST2 and Touch Dreaming~\cite{twist2,dreamtouch} collect demonstrations through teleoperation and learn policies that directly predict robot-level actions, such as joint commands or joint targets.
While this end-to-end formulation provides a direct perception-to-control interface, it couples task-level reasoning with embodiment-specific control, making policy learning sensitive to robot morphology and controller design.

HuMI~\cite{humi} predicts task-space keypoints and uses a low-level controller to generate executable whole-body motions. However, its inverse-kinematics and retargeting process is more tightly integrated with the learned controller, making the intermediate representation less explicit.

In contrast, our framework decomposes whole-body visuomotor learning into three stages. A high-level diffusion policy predicts future keypoint trajectories and gripper widths from visual observations and robot proprioception. An explicit retargeting module maps these predictions to robot-native references, including root pose and joint positions, which a whole-body controller tracks on the physical humanoid. This hierarchy separates task-space planning, kinematic correspondence, and low-level execution, improving interpretability and reducing policy dependence on task-specific control.

\section{Method}

\subsection{Robot-free Data Collection System}

We design a robot-free data collection system for acquiring whole-body humanoid manipulation demonstrations without requiring the physical robot to be present during data collection. The system records human whole-body motion, local wrist-view visual observations, and encoder-measured gripper width in a synchronized and robot-compatible format. The collected demonstrations are then processed for high-level diffusion policy learning.

\begin{figure*}[t]
    \centering
    \includegraphics[width=0.8\textwidth]{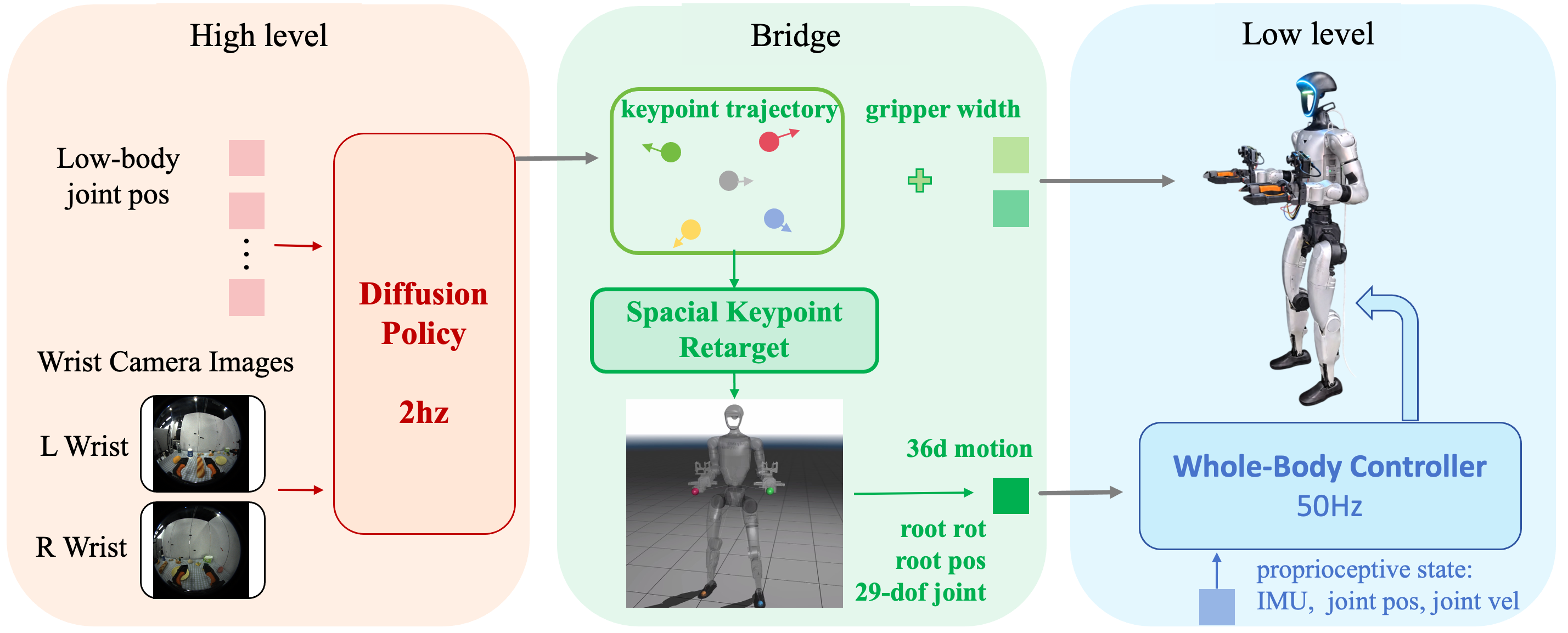}
    \caption{\textbf{BifrostUMI Hierarchical Visuomotor Control.}
    BifrostUMI formulates humanoid visuomotor control as a three-stage hierarchy. 
    A diffusion-based high-level policy predicts task-space keypoint trajectories and gripper commands from wrist-view images and lower-body proprioception. 
    The spatial keypoint retargeting bridge maps these commands into a 36-dimensional robot-native motion representation, including the root pose and joint configurations. 
    A low-level whole-body controller then tracks the retargeted motion with proprioceptive feedback, enabling stable humanoid execution from robot-free demonstrations.}
    \label{method}
\end{figure*}
\textbf{Data-collection hardware:}
As illustrated in Fig.~\ref{data}, the system consists of two main components: a portable VR-based whole-body motion capture system and two handheld manipulation interfaces. The VR component is built on a PICO-based setup, including a headset, handheld controllers, and three lower-body trackers. It provides real-time estimates of the operator's whole-body motion in a portable manner with lightweight calibration.
By fusing inertial, geomagnetic, and active infrared sensing, the PICO tracking pipeline provides stable 6-DoF body-motion estimates in an SMPL-format body representation.
This fused representation improves the reliability of the recorded human motion and facilitates its conversion into transferable keypoint trajectories for humanoid retargeting.
The handheld manipulation interfaces follow the design principles of UMI-style systems and are implemented with a self-developed gripper. Each interface integrates a fisheye camera and a motor-driven rack-and-pinion gripper mechanism, together with a handle that includes a dedicated mounting slot for a PICO controller. This design enables stable grasping while simultaneously capturing wrist-view visual observations and gripper width measurements.

\textbf{Recorded data modalities:}
During robot-free data collection, the operator wears the PICO VR system and holds two handheld grippers attached to VR controllers. Full-body motion is streamed through XRobotoolkit~\cite{xrobotoolkit}. We obtain the 6-DoF poses of the two handheld controllers from the PICO SDK, together with the pelvis, left-foot, and right-foot poses extracted from the SMPL-format body representation. After applying coordinate transformations to align these measurements with the robot keypoint frames, we store them as five whole-body keypoint poses, including the pelvis, two gripper TCPs, and two feet. For lower-body-intensive tasks, we additionally extract the left and right knee poses from the SMPL-format body representation.

To make the collected demonstrations directly usable for humanoid policy learning, we run the proposed Spatial Keypoint Retargeting (SKR) module online during data collection. The human keypoint trajectories are retargeted to robot-native motion references, from which we record the lower-body DoF positions $\mathbf{q}^{\mathrm{low}}_t$. These lower-body states serve as proprioceptive conditioning for the high-level diffusion policy, allowing the policy to predict future keypoint motions conditioned on the support configuration and posture of the robot. In addition, online retargeting enables the operator to visualize the retargeted humanoid motion during collection and check whether the demonstrated motion is mapped to the robot in a natural and kinematically plausible manner. In parallel, the handheld grippers record synchronized wrist-view images using fisheye cameras. The gripper width is measured by the magnetic encoder of the motor drive, providing gripper-state annotations for policy learning.

\textbf{Latency matching.}
For timing-sensitive tasks, we follow the latency-matching procedure of UMI~\cite{umi} to calibrate and compensate for the relative delays among wrist-view observations, gripper-state measurements, high-level policy inference, and robot control.

\subsection{High-Level: Diffusion Policy}

We instantiate the high-level policy as a whole-body extension of Diffusion Policy~\cite{diffusionpolicy}, operating in a sparse task space rather than the full joint space. At each decision step $t$, the policy predicts a receding-horizon action chunk $\mathbf{a}_{t+1:t+H}$ with $H{=}48$ over a default set of five keypoints: the pelvis, left/right TCPs, and left/right feet. The pelvis represents root motion, the feet define the support configuration, and the TCPs specify manipulation endpoints. For tasks requiring substantial knee flexion, we additionally evaluate an augmented seven-keypoint variant that includes the left and right knees.

\begin{figure}[htbp]
\centering
\includegraphics[width=0.40\textwidth]{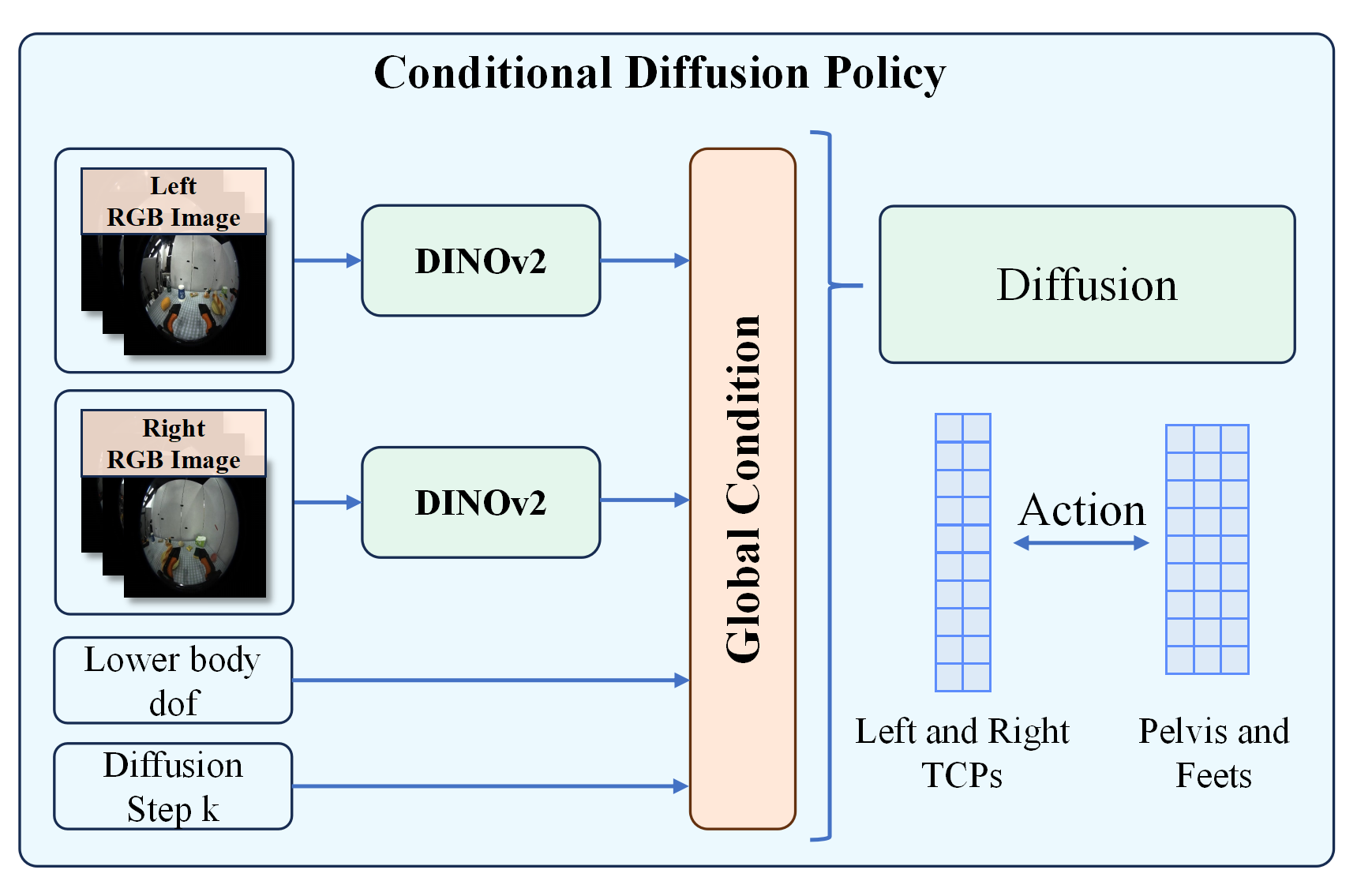}
\caption{
\textbf{Conditional diffusion policy architecture.}
Wrist-view RGB images are encoded by DINOv2 and fused with lower-body proprioception and the diffusion step as the global condition. The diffusion model predicts action trajectories for TCPs and body-support keypoints.
}
\label{fig:policy_architecture}
\end{figure}

\textbf{Action space.}
Each action $\mathbf{a}_\tau \in \mathbb{R}^{47}$ consists of the $6$-DoF poses of five keypoints, represented by a $3$-D translation and a continuous $6$-D rotation~\cite{zhou2019continuity}, together with two scalar gripper widths. This yields $5 \times 9 + 2 = 47$ dimensions for the default configuration. The augmented seven-keypoint variant additionally includes the left and right knee poses, resulting in $7 \times 9 + 2 = 65$ dimensions.

\textbf{Action chunk preparation.}
Given the observation at time $t$, the policy is supervised to predict future keypoint poses over the horizon $\tau=t{+}1,\dots,t{+}H$.
All keypoint poses are expressed in the pelvis frame at the query time $t$. For each keypoint $k$, the future pose is represented relative to its query-time pose:

\begin{equation}
\mathbf{T}^{\mathrm{rel}}_{k,\tau}
=
\bigl(\mathbf{T}^{\mathrm{pel}}_{k,t}\bigr)^{-1}
\mathbf{T}^{\mathrm{pel}}_{k,\tau},
\qquad
\tau=t+1,\dots,t+H .
\label{eq:rel_action}
\end{equation}

Gripper widths remain absolute scalars. Translations and gripper widths are min--max normalized to $[-1,1]$, while $6$-D rotations are left unscaled to preserve valid rotation recovery.

\textbf{Observation.}
The policy is conditioned on one synchronized pair of $224 \times 224$ left/right wrist-view RGB images and a three-frame history of a $15$-D lower-body proprioceptive vector, including $12$ leg joints and $3$ waist joints. Upper-body joints are omitted because the policy expresses manipulation intent through task-space TCP keypoints, with joint configurations resolved downstream by SKR and control. During robot-free data collection, this vector is obtained by applying SKR to the five recorded keypoints and storing the retargeted joint angles. At deployment, the same quantities are read from robot encoders via the Unitree SDK.

\textbf{Inference.}
At runtime, the diffusion policy denoises and inverse-normalizes a relative action chunk. For each predicted keypoint pose, the relative motion is decoded into a pelvis-referenced target by
\begin{equation}
\hat{\mathbf{T}}^{\mathrm{pel}}_{k,\tau}
=
\mathbf{T}^{\mathrm{pel}}_{k,t}
\hat{\mathbf{T}}^{\mathrm{rel}}_{k,\tau},
\qquad
k \in \mathcal{K}, \quad
\tau = t+1,\ldots,t+H.
\label{eq:rel_decode}
\end{equation}
where $\mathcal{K}$ denotes the selected keypoint set and $\mathbf{T}^{\mathrm{pel}}_{k,t}$ is computed from the current robot state through forward kinematics and expressed in the current pelvis frame. The decoded pelvis-referenced targets are then passed to SKR for whole-body retargeting.

\subsection{Bridge: Keypoint Retargeting System}
\label{3c}

\begin{figure}[htbp]
    \centering
    \includegraphics[width=0.45\textwidth]{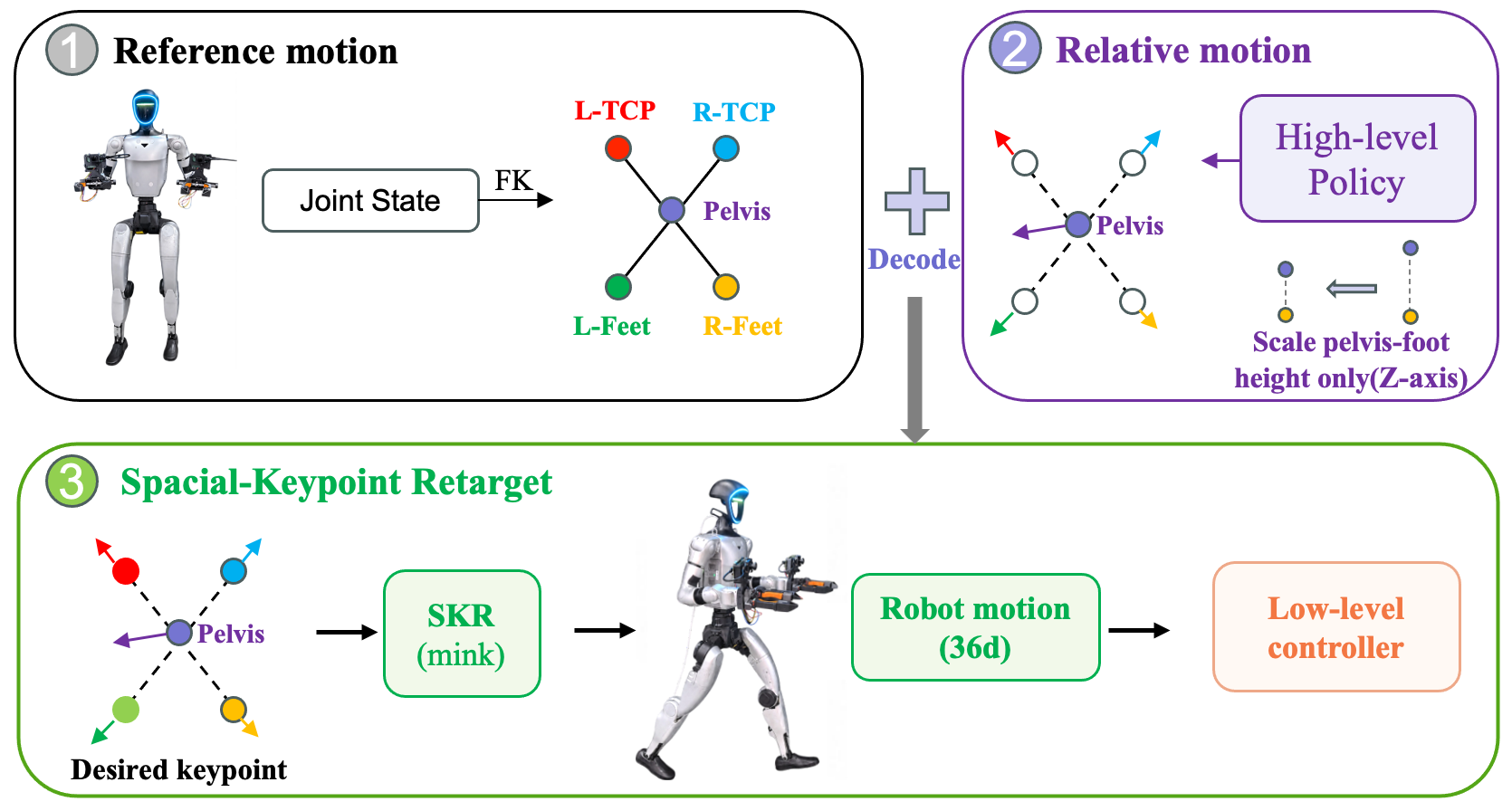}
    \caption{
\textbf{Spatial Keypoint Retargeting (SKR).}
SKR bridges high-level keypoint prediction and low-level whole-body control by converting five task-space keypoints, including the pelvis, two TCPs, and two feet, into robot-native whole-body references. 
Unlike global motion rescaling, SKR preserves metric spatial relationships among the keypoints and only scales the vertical pelvis-to-foot distance to compensate for human--robot height differences. 
The resulting inverse-kinematics solution provides executable joint-level motion commands for the humanoid robot.
}
    \label{skr}
\end{figure}

Human-to-humanoid retargeting is challenging because the demonstrator and the robot differ in body proportions, joint limits, and reachable workspace. Rather than directly transferring human joint motion, we propose \textbf{Spatial Keypoint Retargeting (SKR)}, a morphology-aware pipeline that converts sparse human keypoint trajectories into feasible robot whole-body references. As illustrated in Fig.~\ref{skr}, SKR maps sparse human keypoints to robot-native whole-body references through morphology-aware keypoint adjustment and constrained inverse kinematics.

SKR supports a default five-keypoint layout consisting of the pelvis, two TCPs, and two feet. For lower-body-intensive tasks, two knee keypoints are added:
\begin{equation}
\begin{aligned}
\mathcal{K}_{5}
&=
\{
\text{pelvis},
\text{L/R foot},
\text{L/R TCP}
\}, \\
\mathcal{K}_{7}
&=
\mathcal{K}_{5}
\cup
\{
\text{L/R knee}
\}.
\end{aligned}
\label{eq:keypoint_layout}
\end{equation}

Given the human keypoint poses, SKR first transforms them into the robot coordinate system and aligns the demonstration according to the initial pelvis position. During policy deployment, predicted relative keypoint motions are decoded using the current robot keypoint poses obtained from forward kinematics.

To account for lower-body morphology mismatch while retaining the remaining task-space geometry, SKR performs a pelvis-local anisotropic adjustment only for leg-related keypoints. Let $\mathbf{d}_{i}$ denote the displacement from the pelvis to keypoint $i$. The adjusted displacement is defined as
\begin{equation}
\bar{\mathbf{d}}_{i}
=
\begin{bmatrix}
d_{i,x}\\
d_{i,y}\\
\lambda_i d_{i,z}
\end{bmatrix},
\label{eq:leg_scaling}
\end{equation}
where $\lambda_i=\lambda_{\mathrm{leg}}$ for foot and knee keypoints, and $\lambda_i=1$ for all remaining keypoints. In our experiments, we set $\lambda_{\mathrm{leg}}=0.75$ to account for the leg-length difference between the human demonstrator and the Unitree G1 robot. This adjustment modifies only the pelvis-relative vertical displacement of leg-related keypoints, improving the compatibility of standing and crouching motions with the robot workspace, while leaving all other keypoint coordinates unchanged.

\begin{figure*}[t]
\centering
\makebox[\textwidth][c]{%
\includegraphics[width=0.95\textwidth]{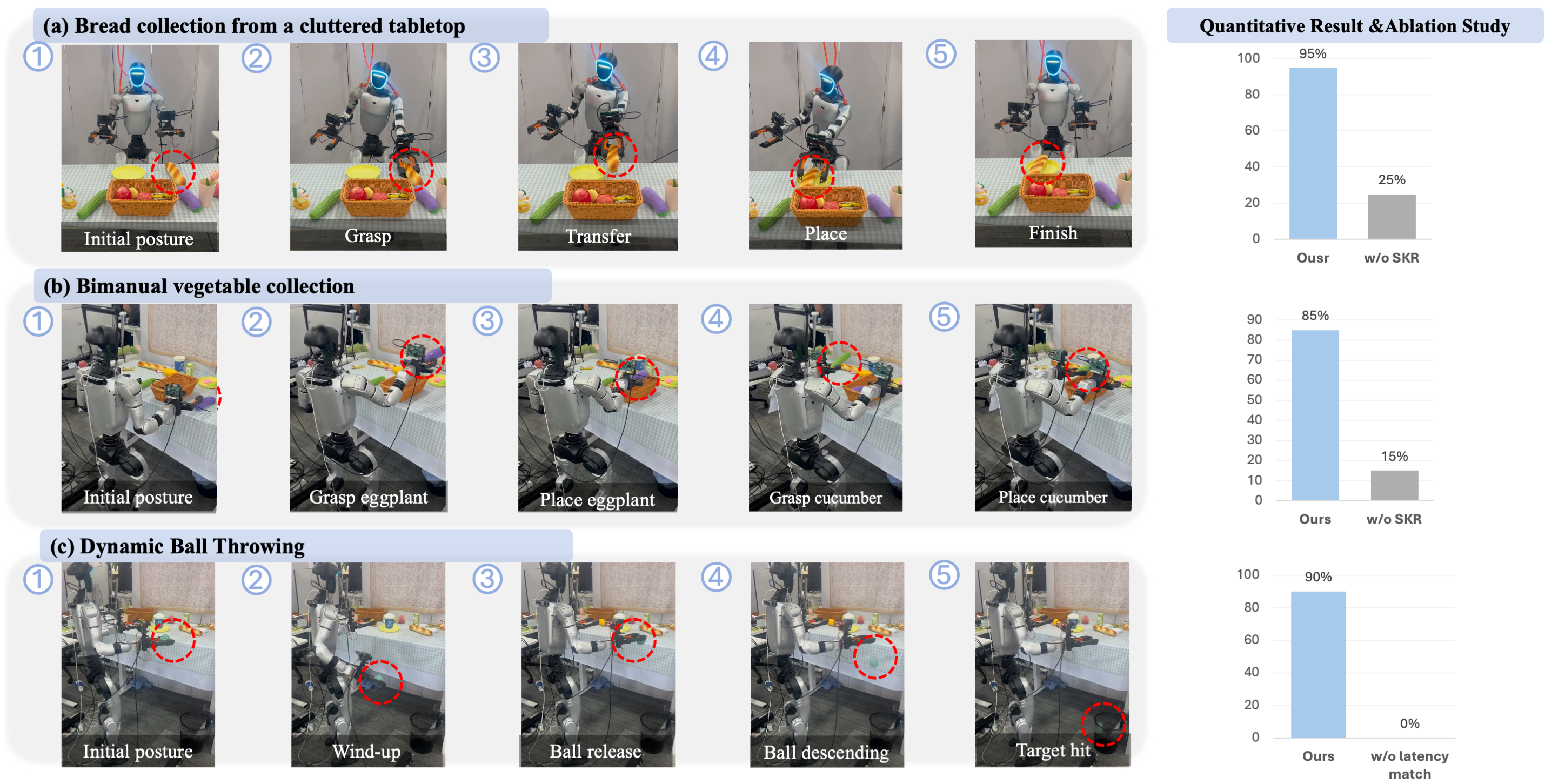}
}
\caption{
\textbf{Real-world evaluation and ablation analysis of BifrostUMI on three humanoid manipulation tasks.}
(a) Cluttered tabletop pick-and-place: the robot grasps bread and places it onto a target plate.
(b) Bimanual vegetable collection: the robot uses the right hand to place an eggplant and the left hand to place a cucumber.
(c) Dynamic ball throwing: the robot performs a wind-up, releases the ball, and hits the target, requiring accurate latency matching.
The left-side bar plots report success rates over 20 independent trials for the corresponding ablations: the first two tasks evaluate the effect of spatial keypoint retargeting (SKR), while the dynamic ball-throwing task evaluates latency matching.
}
\label{fig:real_world_experiments}
\end{figure*}

The adjusted keypoints are associated with semantic robot frames, including the pelvis, toe links, TCPs, and knee links. Fixed calibrated offsets are applied to account for differences between the tracked human keypoints and robot link frames. We then solve a constrained weighted inverse kinematics problem:
\begin{equation}
\mathbf{q}^{*}_{t}
=
\arg\min_{\mathbf{q}_{t}}
\sum_{i \in \mathcal{K}}
w_i
\left\|
\mathbf{e}_{i}(\mathbf{q}_{t})
\right\|_{2}^{2}
+
\lambda_{q}
\left\|
\mathbf{q}_{t}
-
\mathbf{q}_{t-1}
\right\|_{2}^{2},
\label{eq:skr_ik}
\end{equation}
where $\mathbf{e}_{i}$ denotes the pose error between the desired target of keypoint $i$ and the corresponding robot frame computed by forward kinematics. The optimization is subject to robot joint limits and is initialized from the previous solution $\mathbf{q}_{t-1}$ to improve temporal consistency.

In practice, we use a two-stage weighted IK procedure. The first stage prioritizes foot support and orientation consistency, while the second stage refines pelvis and TCP positions. For the seven-keypoint setting, knee-position constraints are additionally activated. The resulting robot root pose and joint references are temporally resampled and passed to the learned whole-body controller for closed-loop execution.

\subsection{Low-Level: Whole-Body Controller}

The low-level controller executes the robot-native full-body references generated by SKR. We adopt a learned motion-tracking controller trained in MJLab~\cite{mjlab} to track these references on the physical Unitree G1 robot.

SKR provides a short-horizon reference motion chunk
\begin{equation}
\mathcal{M}_{\mathrm{ref}}
=
\left\{
\mathbf{p}^{r}_{\tau},
\mathbf{q}^{r}_{\tau},
\mathbf{q}^{j}_{\tau}
\right\}_{\tau=1}^{T},
\end{equation}
where $\mathbf{p}^{r}_{\tau} \in \mathbb{R}^{3}$, $\mathbf{q}^{r}_{\tau} \in \mathbb{R}^{4}$, and $\mathbf{q}^{j}_{\tau} \in \mathbb{R}^{29}$ denote the reference root position, root orientation, and joint configuration, respectively.

The controller runs at 50~Hz and receives the current proprioceptive state together with a temporally resampled reference window. Its 29-dimensional output $\mathbf{a}_{t}$ is interpreted as a residual joint-position action. After clipping and scaling, the desired joint position is computed as
\begin{equation}
\mathbf{q}^{\mathrm{des}}_{t}
=
\mathbf{q}^{0}
+
\mathbf{s} \odot
\mathrm{clip}
\left(
\mathbf{a}_{t}, -a_{\max}, a_{\max}
\right),
\end{equation}
and sent to the robot through joint-space PD control.



\section{Experiments}
\label{sec:experiments}

We evaluate BifrostUMI on real-world humanoid manipulation tasks with a Unitree G1 robot. Our experiments address three central questions:

\begin{itemize}
\item \textbf{Effectiveness of the robot-free data collection and training framework.} Can robot-free demonstrations collected via the VR--UMI interface be transformed into deployable humanoid visuomotor policies?
\item \textbf{Whole-body manipulation capability of BifrostUMI.} Can a sparse keypoint-based representation drive coordinated whole-body behaviors across the hands, feet, and waist?
\item \textbf{Efficiency of robot-free data collection.} Can the robot-free collection interface acquire valid demonstrations more efficiently than teleoperation?

\end{itemize}

\subsection{Effectiveness of the robot-free data collection and training framework}

We first evaluate whether VR--UMI robot-free demonstrations can be transformed into deployable visuomotor policies on a physical humanoid robot. As shown in Fig.~\ref{fig:real_world_experiments}, BifrostUMI is tested on three real-world tasks with a Unitree G1 robot: cluttered tabletop pick-and-place, bimanual vegetable collection, and dynamic ball throwing. These tasks span single-arm visuomotor grasping, coordinated bimanual manipulation, and precisely timed dynamic release.

In cluttered tabletop pick-and-place, the robot localizes bread among distractors, grasps it, and places it onto a target plate, testing wrist-view visual grounding in a cluttered scene. In bimanual vegetable collection, the robot sequentially uses the right hand to place an eggplant and the left hand to place a cucumber, testing whether sparse keypoints can coordinate two arms with distinct object targets while preserving feasible whole-body posture. In dynamic ball throwing, the robot performs a wind-up, releases the ball, and hits the target. Unlike quasi-static manipulation, this task is highly sensitive to release timing and therefore requires latency matching across camera observations, gripper signals, and robot execution.

\begin{figure*}[t]
\centering
\makebox[\textwidth][c]{%
\includegraphics[width=0.95\textwidth]{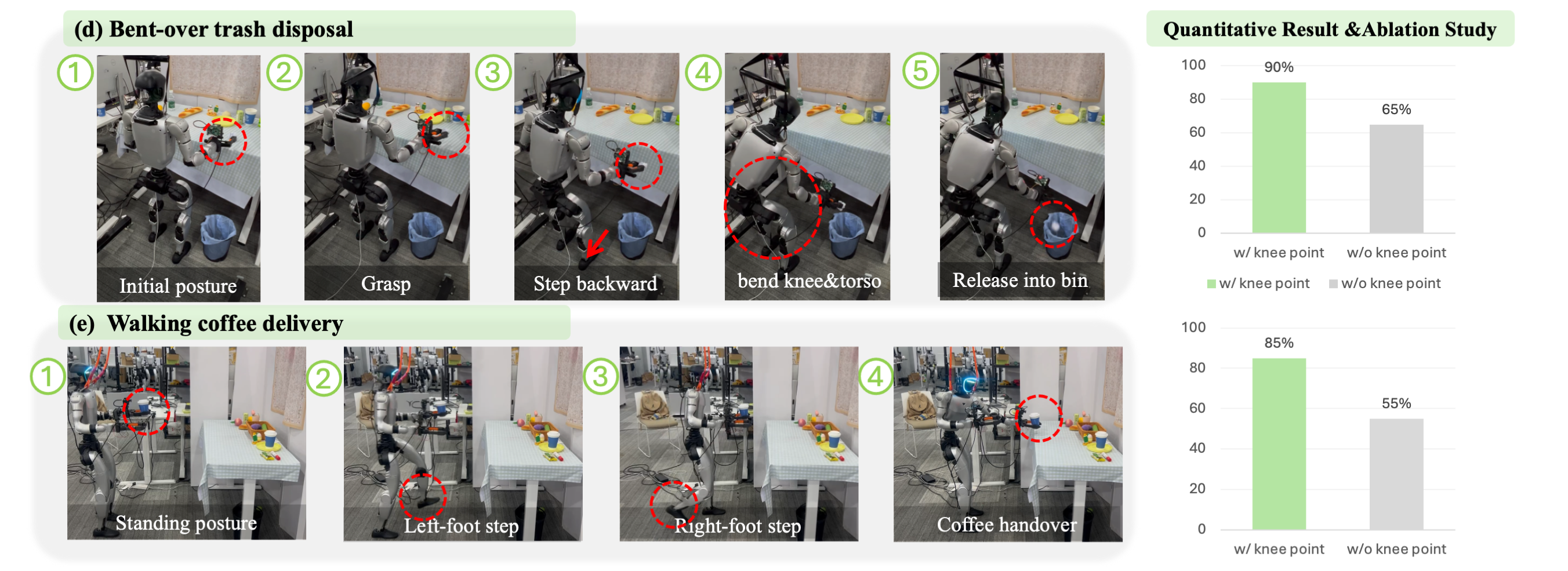}
}
\caption{
\textbf{Real-world evaluation of the augmented seven-keypoint variant of BifrostUMI on two humanoid whole-body manipulation tasks with a Unitree G1 robot.}
(a) Under-table waste disposal: the robot grasps a crumpled paper ball from the tabletop, steps backward, bends its knees and torso, reaches toward a waste bin underneath the table, and releases the object into the bin.
(b) Walking coffee delivery: the robot walks forward with a natural gait, comes to a stable stop, and extends its arm to hand over the coffee.
For these lower-body-intensive tasks, BifrostUMI uses the augmented configuration with two additional knee keypoints.
The left-side bar plots report success rates over 20 independent trials, comparing the augmented seven-keypoint variant with the default five-keypoint configuration for tasks requiring strong waist--leg coordination.
}
\label{fig:real_world_experiments2}
\end{figure*}

The qualitative results in Fig.~\ref{fig:real_world_experiments} show that BifrostUMI transfers robot-free human demonstrations to physical humanoid execution across all three tasks. The robot completes visual localization, grasping, transfer, and placement in the cluttered scene; switches between two arms for sequential bimanual collection; and generates a coordinated wind-up and release motion for dynamic ball throwing, indicating that the learned policy extends beyond slow end-effector reaching.

We further ablate key modules in Fig.~\ref{fig:real_world_experiments}. For the first two tasks, replacing Spatial Keypoint Retargeting with GMR~\cite{gmr} substantially reduces success. This confirms that humanoid deployment requires SKR to perform scale adaptation while preserving the spatial structure of task-relevant keypoints. Without this spatially grounded retargeting, the upper-level visual policy may occasionally recover through closed-loop visual feedback, but such compensation is brittle and unstable. For dynamic ball throwing, removing latency matching also degrades performance, showing that precise synchronization among sensing, gripper state, and robot actuation is critical for timing-sensitive release.

Overall, the complete BifrostUMI pipeline consistently outperforms the ablated variants. These results verify the coupling of its core components: the high-level policy predicts task-relevant sparse keypoint trajectories from robot-free demonstrations, SKR converts human intent into spatially consistent humanoid motion, latency matching enables accurate dynamic execution, and the low-level whole-body controller tracks the retargeted references while maintaining balance during physical interaction.

\subsection{Whole-body manipulation capability of BifrostUMI.}

\label{sec:exp_whole_body_coordination}

We further evaluate the whole-body coordination capability of BifrostUMI on two challenging tasks that require the lower body to participate in manipulation, rather than treating locomotion and manipulation as separate modules. As shown in Fig.~\ref{fig:real_world_experiments2}, we consider under-table waste disposal and walking coffee delivery. The first task requires the robot to grasp a crumpled paper ball from the tabletop, step backward, bend its knees and torso, reach toward a waste bin underneath the table, and release the object into the bin. The second requires the robot to walk forward with a natural gait, stop stably, and extend its arm to deliver a cup of coffee. Together, these tasks test whether BifrostUMI can coordinate whole-body behaviors across the hands, waist, and legs.

Unlike approaches that decouple upper-body manipulation from lower-body locomotion, BifrostUMI represents task intent with sparse whole-body keypoints and executes the retargeted motion using a general-purpose whole-body controller. This design coordinates arm motion with lower-body behaviors such as stepping, torso bending, knee flexion, and balance maintenance. In waste disposal, the robot first maintains an upright support posture while grasping the object, then repositions its feet and lowers its body to bring the hand into the constrained under-table workspace. In coffee delivery, the robot alternates forward steps, stabilizes its stance, and reaches forward to hand over the cup, showing that locomotion and manipulation emerge as a coupled whole-body behavior rather than two independently controlled stages.

The qualitative results in Fig.~\ref{fig:real_world_experiments2} show that BifrostUMI completes both task sequences on the physical Unitree G1 robot. Under-table waste disposal demonstrates that sparse keypoints can encode whole-body posture changes needed to reach targets outside the arm-only workspace, while walking coffee delivery shows that the same framework supports locomotion-manipulation, where lower-body motion directly contributes to task completion rather than merely stabilizing the upper body. The knee-keypoint ablation further shows that, when waist and leg involvement is substantial, knee keypoints are critical for recovering correct lower-body motion, especially knee flexion and body lowering. Without these keypoints, the lower-body posture is underconstrained, leading to less reliable whole-body execution. These results suggest that, combined with a general whole-body controller, BifrostUMI can generate physically executable humanoid skills involving object interaction, stepping, torso bending, knee flexion, and stable reaching.

\subsection{Efficiency of robot-free data collection}

We further evaluate the efficiency of robot-free data collection in BifrostUMI. A scalable humanoid learning framework should reduce not only deployment difficulty but also the effort required to acquire valid training data. We compare BifrostUMI with TWIST2~\cite{twist2}, a state-of-the-art teleoperation-based data collection method, using two operators: an experienced user familiar with both systems and a novice who received only 10 minutes of instruction. For each operator-task pair, we report the number of valid trajectories collected within 10 minutes.

\begin{table}[t]
    \centering
    \caption{
    Valid demonstration throughput of BifrostUMI and TWIST2 within 10 minutes.
    Speedup is computed per operator-task pair.
    }
    \label{tab:data_collection_efficiency}
    \resizebox{\columnwidth}{!}{
    \begin{tabular}{llccc}
        \toprule
        \textbf{Operator} & \textbf{Task} & \textbf{BifrostUMI} & \textbf{TWIST2} & \textbf{Speedup} \\
        \midrule
        \multirow{3}{*}{Novice}
        & Bimanual & 55 & 25 & $2.2\times$ \\
        & Throw trash & 43 & 17 & $2.5\times$ \\
        & Walk + coffee & 61 & 1 & $61.0\times$ \\
        \midrule
        \multirow{3}{*}{Experienced user}
        & Bimanual & 55 & 30 & $1.8\times$ \\
        & Throw trash & 53 & 23 & $2.3\times$ \\
        & Walk + coffee & 62 & 5 & $12.4\times$ \\
        \bottomrule
    \end{tabular}
    }
\end{table}

As shown in Table~\ref{tab:data_collection_efficiency}, BifrostUMI consistently achieves higher valid demonstration throughput than TWIST2 across all operators and tasks, yielding an average speedup of approximately $2.2\times$ on the two non-locomotion tasks and a much larger gain on walking coffee delivery. This gap is especially pronounced for the novice operator, who collects 61 valid walking-delivery demonstrations with BifrostUMI but only one with TWIST2. These results indicate that robot-free collection is particularly effective for loco-manipulation tasks, where direct teleoperation becomes cumbersome and failure-prone.

The operator comparison further suggests a lower operation barrier. With BifrostUMI, the novice achieves throughput close to that of the experienced user across all tasks, whereas TWIST2 shows a larger expertise gap, especially in walking coffee delivery. Thus, BifrostUMI improves data collection efficiency while reducing dependence on operator skill and teleoperation training.

\section{Conclusion}

We introduced BifrostUMI, a robot-free framework for learning humanoid whole-body manipulation from natural human demonstrations. Using a portable VR--UMI interface, BifrostUMI collects wrist-view observations, gripper states, and sparse whole-body keypoint trajectories without the target robot. A hierarchical visuomotor architecture then predicts keypoint motions, retargets them into robot-native whole-body references through SKR, and executes them with a learned whole-body controller. This explicit spatial interface separates task-level visuomotor reasoning from embodiment-specific motion execution.

Experiments on a Unitree G1 demonstrate deployment across cluttered pick-and-place, bimanual collection, dynamic throwing, under-table waste disposal, and walking coffee delivery. These tasks span single-arm and bimanual manipulation, timing-sensitive release, posture adaptation, and locomotion-manipulation. BifrostUMI also achieves higher valid demonstration throughput than teleoperation in our evaluation, supporting robot-free collection as a practical direction for scalable humanoid skill learning.

\bibliographystyle{IEEEtran}
\bibliography{IEEEabrv,mybibfile}
\end{document}